# SUPERVISED CLASSIFICATION PERFORMANCE OF MULTISPECTRAL IMAGES

K Perumal and R Bhaskaran

◆

**Abstract**

Nowadays government and private agencies use remote sensing imagery for a wide range of applications from military applications to farm development. The images may be a panchromatic, multispectral, hyperspectral or even ultraspectral of terra bytes. Remote sensing image classification is one amongst the most significant application worlds for remote sensing. A few number of image classification algorithms have proved good precision in classifying remote sensing data. But, of late, due to the increasing spatiotemporal dimensions of the remote sensing data, traditional classification algorithms have exposed weaknesses necessitating further research in the field of remote sensing image classification..So an efficient classifier is needed to classify the remote sensing imageries to extract information. We are experimenting with both supervised and unsupervised classification. Here we compare the different classification methods and their performances. It is found that Mahalanobis classifier performed the best in our classification.

## 1 INRODUCTION

REMOTE sensing, particularly satellites offer an immense source of data for studying spatial and temporal variability of the environmental parameters. Remotely sensed imagery can be made use of in a number of applications, encompassing reconnaissance, creation of mapping products for military and civil applications, evaluation of environmental damage, monitoring of land use, radiation monitoring, urban planning, growth regulation, soil assessment, and crop yield appraisal [1]. Generally, remote sensing offers imperative coverage, mapping and classification of land-cover features, namely vegetation, soil, water and forests. A principal application of remotely sensed data is to create a classification map of the identifiable or meaningful features or classes of land cover types in a scene [2]. Therefore, the principal product is a thematic map with themes like land use, geology and vegetation types [3]. Researches on image classification based remote sensing have long attracted the interest of the remote sensing community since most environmental and socioeconomic applications are based on the classification results [4].

- K.*Perumal is with Department of Computer Science, D.D.E., Madurai Kamaraj University, Madurai-625021, Tamil Nadu, India.*
- R.*Bhaskaran is with School of Mathematics, Madurai Kamaraj University, Madurai-625021, Tamil Nadu, India.*

Remote sensing image classification can be viewed as a joint venture of both image processing and classification techniques. Generally, image classification, in the field of remote sensing is the process of assigning pixels or the basic units of an image to classes. It is likely to assemble groups of identical pixels found in remotely sensed data into classes that match the informational categories of user interest by comparing pixels to one another and to those of known identity .

Several methods of image classification exist and a number of fields apart from remote sensing like image analysis and pattern recognition make use of a significant concept, classification. In some cases, the classification itself may form the entity of the analysis and serve as the ultimate product. In other cases, the classification can serve only as an intermediate step in more intricate analyses, such as land degradation studies, process studies, landscape modeling, coastal zone management, resource management and other environment monitoring applications. As a result, image classification has emerged as a significant tool for investigating digital images. Moreover, the selection of the appropriate classification technique to be employed can have a considerable upshot on the results of whether the classification is used as an ultimate product or as one of numerous analytical procedures applied for deriving information from an image for additional analyses [5].

For the purpose of classification and mapping of vegetation over large spatial scales remotely sensed data are generally used. This acts as a substitute for traditional classification methods, which necessitates expensive and time-intensive field surveys [7]. The multispectral airborne as well as satellite remote sensing technologies have been utilized as a widespread source for the purpose of remote classification of vegetation [6] ever since the early 1960s. Owing to the development of airborne and satellite hyperspectral sensor technologies, the limitations of multispectral sensors [9] have been overwhelmed in the past two decades. Hyperspectral remote sensing imagers obtain several, very narrow, contiguous spectral bands all through the visible, near-infrared, mid-infrared, and thermal infrared portions of the electromagnetic spectrum [10]. Hyperspectral images have taken a significant part in extensive applications of water resource management, agriculture and environmental monitoring [8]. A widespread research and development has been performed in the field of hyperspectral remote sensing.

Advanced classification technologies and large quantities of Remotely Sensed Imagery [11] provide opportunity for useful results. Extracting interesting patterns and rules from data sets composed of images and associated ground data is very important for resource discovery. Image classification is an important part of the remote sensing data mining. The performance [12] of the classifiers depends upon the data. So a better understanding of data is necessary for further advances. Such an understanding is not possible in the traditional theoretical studies. Comparative studies of classifiers that relate their performances to data characteristics have received attention only recently. Successful classification requires experience and experimentation. The analyst must select a classification method that will best accomplish a specific task. At present it is not possible to state which classifier is best for all situation as the characteristics of each image and the circumstances for each study vary so greatly.

## 2 SUPERVISED CLASSIFICATION

Image classification in the field of remote sensing, is the process of assigning pixels or the basic units of an image to classes. It is likely to assemble groups of identical pixels found in remotely sensed data, into classes that match the informational categories of user interest by comparing pixels to one another and to those of known identity [3]. Several methods of image classification exist and a number of fields apart from remote sensing like image analysis and pattern recognition make use of a significant concept, classification. In some cases, the classification itself may form the entity of the analysis and serve as the ultimate product. In other cases, the classification can serve only as an intermediate step in more intricate analyses, such as land-degradation studies, process studies, landscape modeling, coastal zone management, resource management and other environment monitoring applications. As a result, image classification has emerged as a significant tool for investigating digital images. Moreover, the selection of the appropriate classification technique to employ can have considerable upshot on the results, of whether the classification is used as an ultimate product or as one of numerous analytical procedures applied for deriving information from an image for additional analyses [13].

The remote sensing literature presents with a number of supervised methods that have been developed to tackle the multispectral data classification problem. The statistical method employed for the earlier studies of land-cover classification is the maximum likelihood classifier. In recent times, various studies have applied artificial intelligence techniques as substitutes to remotely-sensed image classification applications. In addition, diverse ensemble classification method has been proposed to significantly improve classification accuracy [12]. Scientists and practitioners have made great efforts in developing efficient classification approaches and techniques for improving classification accuracy.

The quality of a supervised classification [3] depends on the quality of the training sites. All the supervised classifications usually have a sequence of operations that must be followed. 1. Defining of the Training Sites. 2. Extraction of Signatures. 3. Classification of the Image. The training sites are done with digitized features. Usually two or three training sites are selected. The more training site is selected, the better results can be gained. This procedure assures both the accuracy of classification and the true interpretation of the results. After the training site areas are digitized then the statistical characterizations of the information are created. These are called signatures. Finally the classification methods are applied.

## 3 MATERIALS AND METHODS

In this research, we have made use of land cover images obtained from remote sensing for experimentation. The Indian Remote Sensing Satellite IRS-p6 Liss3 (Indian Remote Sensing – Linear Imaging Self-Scanning Sensor 3) multispectral sensor operating in bands R, G, B, NIR and SWIR with a swath of 141 km. as in Fig.1 has been used. The dataset consists of 2282 x 2507 pixels and covers Madurai, Tamil Nadu. The advantage of using this dataset is the availability of the referenced image produced from field survey, which is used for the accuracy purpose In this research, we have made use of land cover images obtained from remote sensing for experimentation. The Indian Remote Sensing Satellite IRS - P6 LISS-3 (Indian Remote Sensing-Linear Imaging Self-Scanning Sensor 3) data having values in bands 2, 3, 4 and 5 in the format of LGSOWG ((Landsat Ground Station Operators Working Group) or Super Structure Format) of the electromagnetic spectrum with a spatial resolution procured from DPSD/SIPG/SIIPA, ISRO, Ahmedabad, INDIA was used as the test input in the proposed research. Land cover images chosen for study correspond to the Madurai district of TamilNadu State, India. The land use image encompasses of large areas of farmland, residents, wetlands, some of the water bodies and more. The original remote sensing image in false colors with RGB: 432 with its characteristics Table 1is shown in Fig. 1

A multispectral image covers enormous areas of land cover and is inherently difficult to process on this entire multispectral image. The bare soil sites shows up as white areas because of sloping of the reflectance distributions at these bands. The red areas indicate vegetation, which reflects more in the near infrared than in green visible wavelengths. The river shows up as dark area because near infrared wavelengths are absorbed in the very upper surface of water. A ground truth image (reference image) is generated by field study campaign as in Fig. 2. Random sampling is carried out to select the pixels for training and testing the classifiers.

Table 1 Characteristics of Liss-3

```
PRODUCT 1 :
Product number                     : JobId
   (Twelve Character)
Satellite ID                       : P6
   (Two Character)
Sensor                             : L-3
         (Three Character)
Path-Row                           : 096-055
      (Seven Character)
Date& time of Acquisition          : 28-MAR-02
05:50:39   Eighteen Character)
Product Code                       : STPCD027J
      (Nine Character)
Orbit Number                       : 1
Image Layout                       :BSQ
Number of Bands                    : 4
Bands Present in Product           : 2 3 4 5
Bands in this volume               : 2 3 4 5
File Header                        : 540
Line Header (Prefix Bytes )        : 32
Line Trailer (Suffix Bytes )       : 0
Scan Lines                         : 5545
Pixels                             : 5918
Bytes Per Pixel                    : 1
Image Record Length (Bytes)        : 5950
No of Volume                       : 1/1
 (3 Chracter-CurVol/NoOfPhyVol)
```

## 4 IMPLEMENTATION

The main aim of the study is to evaluate the performance of the different classification algorithms using the multispectral data. This is implemented with ENVI 4.2 [14]. In a similar way, the classification algorithms can be applied for the hyperspectral data [15].

### 4.1 Parallelepiped Classifier

It is a very simple supervised classifier. Here two image bands are used to determine the training area of the pixels in each band based on maximum and minimum pixel values. Although parallelepiped is the most accurate of the classification techniques, it is not most widely used. It leaves many unclassified pixels and also can have overlap between training pixels. The data values of the candidate pixel are compared to upper and lower limits. The classified image is shown in Fig. 3.

### 4.2 Minimum Distance Technique

It is based on the minimum distance decision rule that calculates the spectral distance between the measurement vector for the candidate pixel and the mean vector for each sample. Then it assigns the candidate pixel to the class having the minimum spectral distance. The classified image is shown in Fig. 4.

### 4.3 Maximum Likelihood

This Classification uses the training data by means of estimating means and variances of the classes, which are used to estimate probabilities and also consider the variability of brightness values in each class. This classifier is based on Bayesian probability theory. It is the most powerful classification methods when accurate training data is provided and one of the most widely used algorithm. The classified image is shown in Fig. 5.

### 4.4 Spectral Angle Mapper

The Spectral Angle Mapper is a physically based spectral classification that uses an n-dimensional angle to match pixels to reference spectra. This algorithm determines the spectral similarity between two spectra by calculating the angle between the spectra, treating them as vectors in a space with dimensionality equal to number of bands. The classified image is shown in Fig. 6.

### 4.5 Artificial Neural Network (ANN) Classifier

A multi-layered feed-forward ANN [16] is used to perform a non-linear classification. This model consists of one input layer, at least one hidden layer and one output layer and uses standard back propagation for supervised learning. Learning occurs by adjusting the weights in the node to minimize the difference between the output node activation and the output. The error is back propagated through the network and weight adjustment is made using a recursive method. The classified image is shown in Fig. 7

### 4.6 Mahalanobis distance

Mahalanobis distance classification is similar to minimum distance classification except that the covariance matrix is used. The Mahalanobis distance algorithm assumes that the histograms of the bands have normal distributions. The classified image is shown in Fig. 8.

A study of the performance of various classifiers mentioned above based on the overall accuracy, kappa coefficient, and confusion matrix as in table 2 and 3 is made. It is observed that Mahalanobis classification method is determined to be the most accurate. One of the reasons is it filters out shadows and also it classifies the highly varied clusters. Overall Accuracy = (67903/68047) = 99.7884% and Kappa Coefficient = 0.9716

The error matrix shows the accuracy of Mahalanobis classification method using the image provided. Similarly the error matrices for other classification were found out.

The minimum distance classifier is found to be the least accurate with the lowest accuracies. The output of the classification is shown in figure [Fig. 3-8]. Overall, the Mahalanobis classifier shows the highest accuracy assessment for this particular area

## 5 PERFORMANCE

Table 2 Mahalanobis Classification Error matrix Ground Truth (Pixels)

| Class | Class1 | Class2 | Class3 | Total |
|---|---|---|---|---|
| Unclassified | 136 | 5 | 2 | 143 |
| Class1 [Red] | 65367 | 1 | 0 | 65368 |
| Class2 [Green] | 0 | 1514 | 0 | 1514 |
| Class3 [Blue] | 0 | 0 | 1022 | 1022 |
| Total | 65503 | 1520 | 1024 | 6047 |

Table 3 Mahalanobis Classification Error matrix Ground Truth (Percent)

| Class | Class | Class2 | Class3 | Total |
|---|---|---|---|---|
| Unclassified | 0.21 | 0.33 | 0.20 | 0.21 |
| Class1 [Red] | 99.79 | 0.07 | 0.00 | 96.06 |
| Class2 [Green] | 0.00 | 99.61 | 0.00 | 2.22 |
| Class3 [Blue] | 0.00 | 0.00 | 99.80 | 1.50 |
| Total | 100.00 | 100.00 | 100.00 | 100.00 |

## 6 CONCLUSIONS

In this paper we have compared the performance of various classifiers and found that the Mahalanobis classifier outperforms even advanced classifiers. This accurate but simple classifier shows the importance of considering the data set - classifier relationship for successful image classification. Further studies are required to improve the use of classifiers to increase the applicability of such methods.

There is a need to develop new algorithms to classify more number of classes and more land cover and land use. This would save more human resources, time and fund.

## REFERENCES

[1] James A. Shine and Daniel B. Carr, "A Comparison of Classification Methods for Large Imagery Data Sets", *JSM 2002 Statistics in an ERA of Technological Change-Statistical computing section*, New York City, pp.3205-3207, 11-15 August 2002.

[2] Jasinski, M. F., "Estimation of subpixel vegetation density of natural regions using satellite multispectral imagery", *IEEE Trans. Geosci. Remote Sensing*, Vol. 34, pp. 804–813, 1996.

[3] C. Palaniswami, A. K. Upadhyay and H. P. Maheswarappa, "Spectral mixture analysis for subpixel classification of coconut", *Current Science*, Vol. 91, No. 12, pp. 1706 -1711, 25 December 2006.

[4] D. Lu, Q. Weng, "A survey of image classification methods and techniques for improving classification performance", *International Journal of Remote Sensing*, Vol. 28, No. 5, pp. 823-870, January 2007.

[6] Landgrebe D., "On information extraction principles for hyperspectral data", *Cybernetics 28* part c, Vol. 1, pp. 1-7, 1999.

[7] M. Govender, K. Chetty, V. Naiken and H. Bulcock, "A comparison of satellite hyperspectral and multispectral remote sensing imagery for improved classification and mapping of vegetation", *Water SA, Vol. 34*, No. 2, April 2008.

[8] Smith R. B., "Introduction to hyperspectral imaging", 2001a. www.microimages.com (Accessed 11/03/2006).

[9] Smith R. B., "Introduction to remote sensing of the environment", 2001b. **www.microimages.com .**

[10] M. Govender, K. Chetty and H. Bulcock, "A review of hyperspectral remote sensing and its application in vegetation and water resource studies", *Water SA, Vol. 33,* No. 2, pp.145-151, April 2007.

[11] J. A. Richards., Remote *Sensing Digital Image Analysis, Springer-Verlag*, Berlin, pp. 240-255, (1999).

[12] D. Landgrebe., Hyperspectral image data analysis, *IEEE signal process*. Mag., volume 19, pp. 17-28, (Jan 2002).

[13] B. Gabrya, L. Petrakieva., Combining labeled and unlabelled data in the design of pattern classification systems, *International Journal of Approximate Reasoning*, (2004).

[14]**www.envi.com.br/asterdtm/english/download/asterdtm.html.**

[15] X. H. Liu, A. K. Skidmore, V. H. Oosten., Integration of classification Methods for improvement of land-cover map accuracy, *ISPRS Journal of Photogrammetry & Remote Sensing*, 64, pp. 1189-1200, (2002).

[16] P. M. Atkinson, A. R. L. Tatnall., *Introduction to neural networks in remote sensing, International Journal of Remote sensing, volume* 11, pp. 699-709, (1997).

**K. Perumal** received the B.Sc and M.Sc from the Madurai Kamaraj University in 1986 and 1988 respectively and M. Phil from Manomaniam Sundaranar University Tirunelveli.. He then joined Madurai Kamaraj University and currently working as Associate Professor of Computer Science. He is currently pursuing the Ph.D degree, working closely with Prof. R.Bhaskaran. He works in the field of image processing, pattern recognition, computer vision and data mining.

**R. Bhaskaran** got his M.Sc from IIT Chennai in 1973 and received the Ph.D from the Madras University in 1980. He then joined as Lecturer in Madurai Kamaraj University and currently working as senior professor of Mathematics. His area of interest includes non-Archimedean functional nalysis,graph theory related to computer science, Lindenmayer systems, computer applications (image rocessing, datamining and software development for learning mathematics).

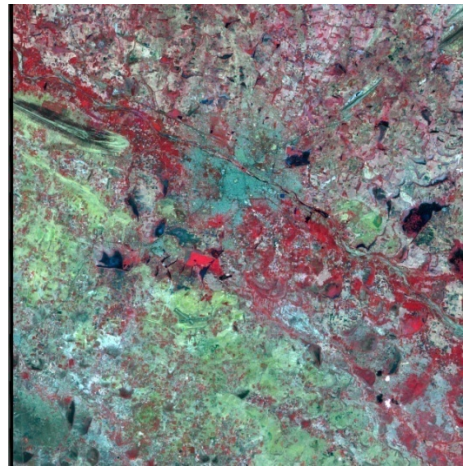

Fig. 1 IRS-1D imagery

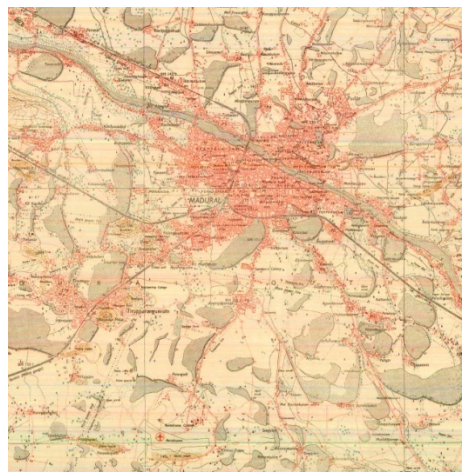

Fig. 2 Madurai TopoSheet

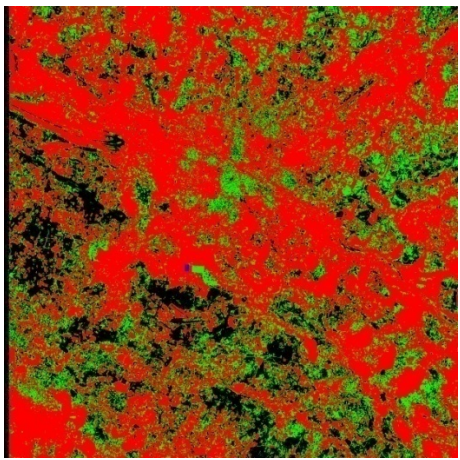

Fig. 3 Parallelepiped Classification

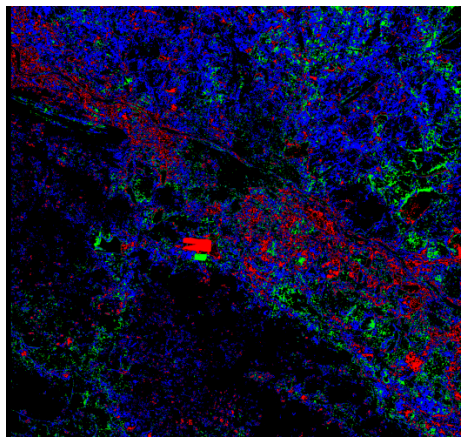

Fig. 6 Spectral Angle Mapper classification

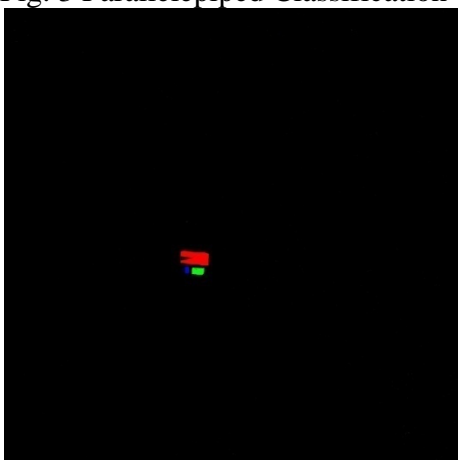

Fig. 4 Minimum Liklihood classification

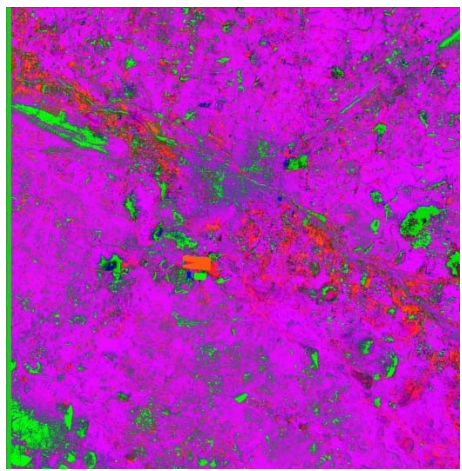

Fig. 7 Neural Network Classification

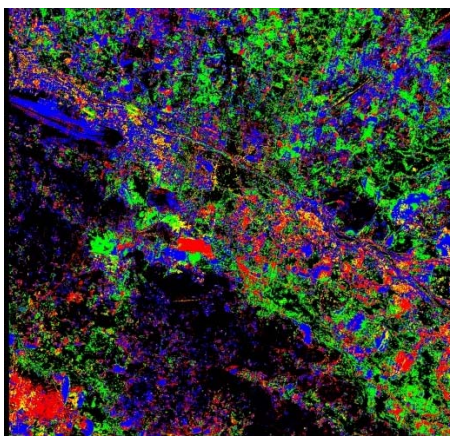

**Fig. 5** Maximum likelihood classification

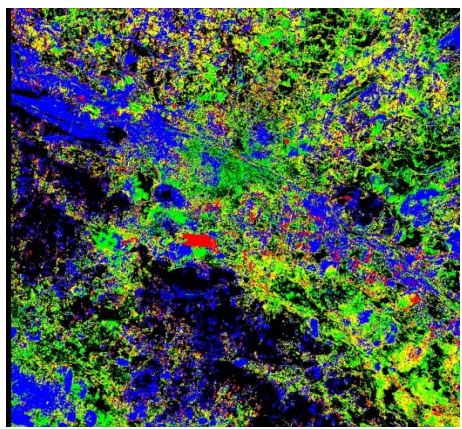

Fig. 8 Mahalanobis Classification